\documentclass[letterpaper,10pt,conference]{ieeeconf}

\IEEEoverridecommandlockouts
\usepackage{amsmath}
\usepackage{amssymb}
\usepackage{booktabs}
\usepackage{cite}
\usepackage{float}
\usepackage{graphicx}
\usepackage{multirow}
\usepackage{placeins}
\usepackage{xcolor}
\usepackage{tikz}
\usepackage{pgfplots}
\usetikzlibrary{fit}
\pgfplotsset{compat=1.18}

\graphicspath{{figures/}}

\title{\LARGE \bf ViTacWorld: Scaling Visuo-Tactile World Models for Contact-Rich Robot Manipulation}
\author{Yunao Huang$^{1,2}$, Shiyu Sang$^{1,2}$, Suting Ni$^{1}$, Haotao Lu$^{1}$,\\
Shijie Wu$^{1}$, Rui Lv$^{1}$, Ziyang Guo$^{1}$, Ye Shi$^{1,2}$, Jingya Wang$^{1,*}$%
\thanks{$^{1}$ShanghaiTech University. $^{2}$InstAdapt.}%
\thanks{\raggedright $^{*}$Corresponding author: Jingya Wang (wangjingya@shanghaitech.edu.cn).}%
\thanks{\raggedright Emails: \{huangya2024, sangshy2025, nist2024, luht2025, wushj12023, lvrui2026, guozy12025, shiye, wangjingya\}@shanghaitech.edu.cn.}%
}

\usepackage[hidelinks]{hyperref}
\hypersetup{
  pdftitle={ViTacWorld: Scaling Visuo-Tactile World Models for Contact-Rich Robot Manipulation},
  pdfauthor={Yunao Huang, Shiyu Sang, Suting Ni, Haotao Lu, Shijie Wu, Rui Lv, Ziyang Guo, Ye Shi, Jingya Wang},
  pdfsubject={Preprint},
  pdfkeywords={Visuo-Tactile World Models, Contact-Rich Manipulation, Policy Improvement, Policy Evaluation}
}

\begin{document}
\maketitle


\begin{abstract}
Contact-rich robot manipulation requires physical interaction cues that are often invisible to cameras, making tactile sensing essential for robust control. However, scaling visuo-tactile robot learning remains difficult because real tactile interaction data are expensive to collect, hardware-dependent, and limited in task and scene diversity. We present ViTacWorld, an action-conditioned visuo-tactile world model for scalable contact-rich robot manipulation. ViTacWorld leverages public real tactile datasets and a constructed simulation environment to scale visuo-tactile-action data, exploiting the fact that tactile signals are directly grounded in physical contact and can exhibit a smaller simulation-to-real gap than purely visual observations. The model is first pretrained with large-scale real and simulated visuo-tactile trajectories, and then finetuned with real-world policy rollouts to better match downstream manipulation behaviors. Given robot actions, ViTacWorld predicts temporally aligned visual observations and tactile feedback, enabling visuo-tactile-action rollout generation. To the best of our knowledge, ViTacWorld is the first framework that uses a world model for robot visuo-tactile-action trajectory generation and policy evaluation. It serves two roles: synthesizing rollouts to improve downstream tactile policies, and evaluating policies by predicting action-conditioned visuo-tactile outcomes under controlled action sequences. Experiments on contact-rich manipulation tasks show that ViTacWorld generates physically meaningful rollouts, improves policy performance through scalable data augmentation, and enables action-conditioned policy evaluation. 

Project page: \url{https://vitacworld.github.io/}
\end{abstract}



\section{Introduction}
\label{sec:introduction}


Contact-rich manipulation tasks such as insertion, peeling, and plugging are central to real-world robot manipulation, yet difficult to accomplish with vision alone due to occlusion, constrained interfaces, and subtle force-dependent dynamics. Tactile sensing directly captures physical interaction at contact, making it a key modality rather than a supplementary signal. However, scaling visuo-tactile robot learning remains challenging because task-relevant visuo-tactile-action trajectories are difficult to obtain at scale. Real-robot teleoperation with tactile sensors provides faithful data, but it is expensive and slow: vision-based tactile sensors such as GelSight-style devices~\cite{dong2017improvedgelsight,yuan2017gelsight} are costly, have limited lifetime under repeated contact, and contact-rich demonstrations are often hard to collect reliably.

Simulation offers a scalable alternative. Recent systems such as TacSL~\cite{akinola2025tacsl}, TacEx~\cite{nguyen2024tacex}, and UniVTAC~\cite{chen2026univtac} enable visuo-tactile sensor simulation and simulated contact-rich manipulation data collection in physics environments. Compared with purely visual observations, tactile signals are more directly grounded in local contact geometry and force response, which makes simulated tactile feedback a particularly promising source for scaling visuo-tactile data. Nevertheless, simulated contact-rich data remain constrained by task diversity, asset fidelity, contact modeling, and the sim-to-real gap; many everyday tasks involving deformable objects, frictional surface interaction, wiping, peeling, or precise real-world fixtures remain difficult to reproduce faithfully. UMI-style robot-free teaching systems~\cite{chi2024umi,zhaxizhuoma2025fastumi} decouple demonstration collection from full robot teleoperation, and recent tactile extensions such as FreeTacMan~\cite{wu2025freetacman} and exUMI~\cite{xu2025exumi} further enable visuo-tactile data collection; however, they still require dedicated tactile hardware and may introduce embodiment or viewpoint gaps. These limitations motivate a broader question: can we scale visuo-tactile robot learning by combining public tactile data, simulation-generated interaction data, and real-world policy rollouts within a unified action-conditioned world model? 

\begin{figure}[!t]
    \centering
    \includegraphics[width=\columnwidth]{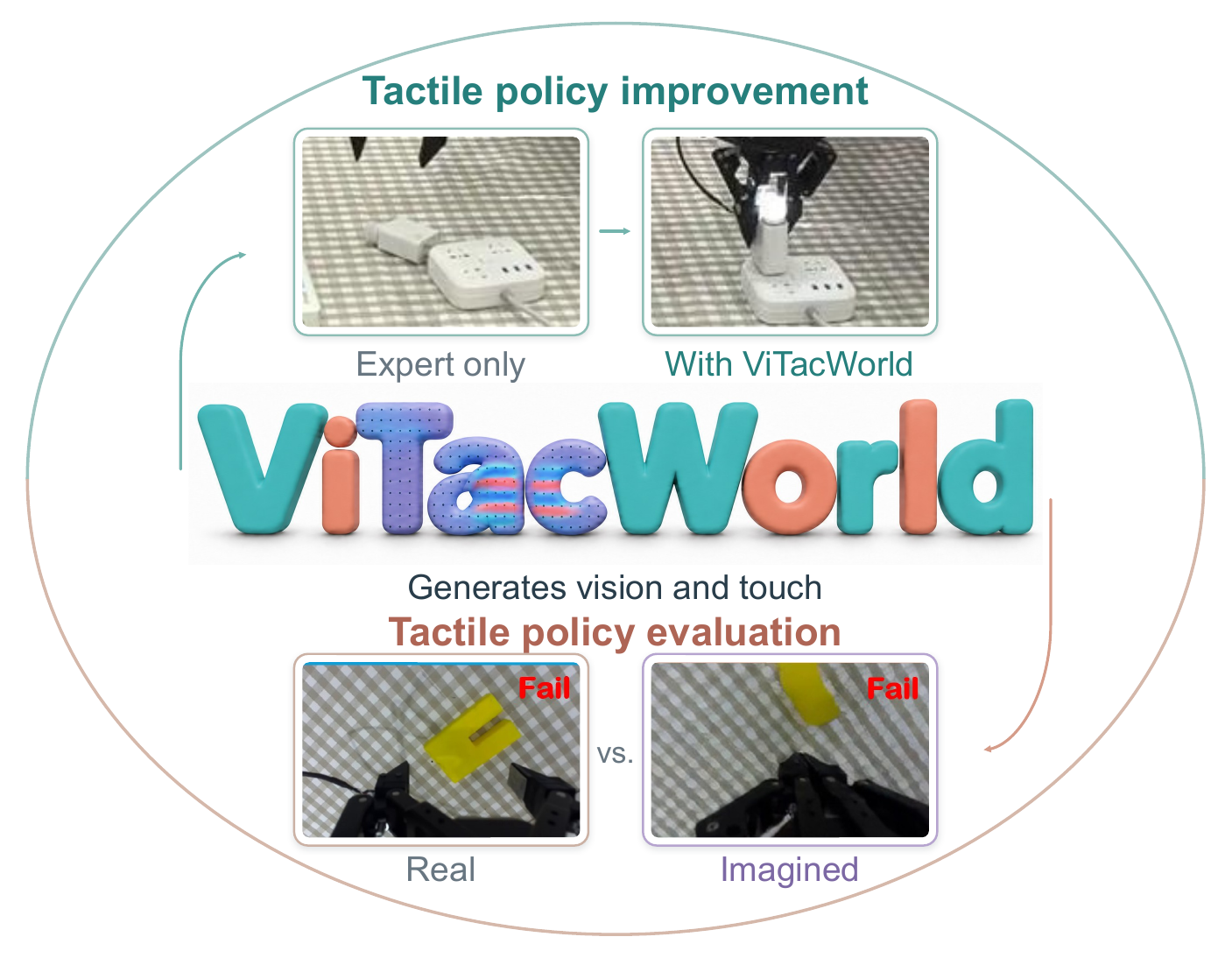}
    \caption{\textbf{ViTacWorld} generates action-conditioned visual and tactile observations for tactile policy improvement and evaluation. Top: generated rollouts improve charger plugging compared with expert-only training. Bottom: real and imagined U-Block insertion examples exhibit the same failure outcome.}
    \label{fig:teaser}
\end{figure}

Recent progress in video foundation models~\cite{blattmann2023stablevideo,chi2025wow} suggests a new route for scaling robot data. General video generators such as Wan~\cite{wan2025wan} demonstrate strong temporal generation ability, while Physical-AI-oriented world models such as Cosmos~\cite{ali2025cosmospredict25} further adapt this capability to robotics, simulation, and dream data generation. In particular, action-conditioned robot world~\cite{ali2025cosmospredict25,huang2025particleformer} models provide a useful prior for modeling how robot actions change future observations, making them a natural foundation for generating manipulation rollouts. Our insight is to extend this action-conditioned world-modeling capability beyond vision: by learning how robot actions jointly evolve visual observations and tactile feedback, a world model can synthesize contact-rich visuo-tactile-action trajectories and predict policy outcomes under controlled action sequences. This perspective differs from using world models only as video predictors, planners \cite{higuera2026visuo}, or internal components of end-to-end world-action policies \cite{zheng2026omnivta,yuan2026vtam}. We instead study visuo-tactile world models as reusable data generators for downstream tactile policies.

We propose ViTacWorld (Fig.~\ref{fig:teaser}), a visuo-tactile world-modeling framework that generates contact-rich manipulation rollouts with temporally aligned visual and tactile observations conditioned on robot action chunks. 
Rather than learning motion and tactile dynamics jointly from scratch, ViTacWorld inherits strong motion priors and focuses its capacity on acquiring tactile dynamics at low adaptation cost. Architecturally, we model tactile sensing as an additional generated view and align the visual and tactile streams via view-aware conditioning and cross-view attention, which ensures cross-view consistency between the two modalities. To enhance generalization and generation quality, we first pretrain the model on large-scale public and simulated data, then finetune it on real task demonstrations and policy rollouts from our target setup. Notably, the simulated data, constructed via real-to-sim techniques, serves as a complementary source during pretraining, helping the model better inherit knowledge from large-scale data while adapting to our target tasks.
We validate the generated data by finetuning representative downstream tactile policies, including lightweight imitation policies and VLA-style policies that consume visual and tactile observations. Improved contact-rich manipulation performance after finetuning indicates that the generated rollouts are not only visually plausible, but useful as dream visuo-tactile data for policy learning.


In summary, our contributions are: \textbf{(i)} we propose ViTacWorld, an action-conditioned visuo-tactile world-modeling framework that enables temporally aligned visuo-tactile-action rollout generation for contact-rich robot manipulation; to the best of our knowledge, this is the first world-model-based framework for robot visuo-tactile-action trajectory generation; \textbf{(ii)} we develop a scalable training pipeline that combines public visuo-tactile-action data, simulation-generated visuo-tactile interactions, and real-world policy rollout finetuning to scale visuo-tactile world modeling; and \textbf{(iii)} we show that ViTacWorld not only improves downstream tactile policies through synthetic rollout augmentation, but also supports action-conditioned policy evaluation by predicting visuo-tactile outcomes under controlled action sequences. 

\section{Related Work}
\label{sec:related_work}

\paragraph{Visuo-Tactile Policies for Contact-Rich Manipulation.}
Tactile sensing has been widely studied as a complementary modality to vision for contact-rich robot manipulation. 
Recent work integrates tactile feedback into end-to-end manipulation policies, including visuo-tactile imitation or diffusion policies~\cite{chi2025diffusion, ze20243d, wu2025afforddp} such as 3D-ViTac~\cite{Huang20243DViTacLF}, Reactive Diffusion Policy~\cite{xue2025reactive}, and PolyTouch~\cite{zhao2025polytouch}, among others, which improve fine-grained manipulation, grasp adjustment, and contact-rich control under visual ambiguity.
In parallel, tactile-augmented and force-aware VLA models incorporate tactile or force feedback into generalist robot policies to improve contact-sensitive decision making, including VLA-Touch~\cite{bi2025vlatouch}, Tactile-VLA~\cite{huang2025tactilevla}, ForceVLA~\cite{yu2025forcevla}, and related variants~\cite{li2026forcevla2,hao2025tla,morissette2026tactile,zhang2025vtla}, among others.
Recent visuo-tactile world-action models, including DreamTacVLA~\cite{ye2026dreamtacvla}, OmniVTA~\cite{zheng2026omnivta}, VTAM~\cite{yuan2026vtam}, and VT-WAM~\cite{tian2026vtwam}, integrate tactile prediction with action generation. Complementary predictive policies, such as TacForeSight~\cite{zang2026tacforesight} and TouchWorld~\cite{zhou2026touchworld}, use anticipated tactile states to guide contact-rich control. 
Together, these works demonstrate that tactile feedback provides policy-relevant information that cannot be reliably inferred from vision alone. 
However, they primarily focus on how policies consume tactile observations or integrate tactile prediction into the policy itself, whereas our work focuses on the complementary question of how to generate additional task-relevant visuo-tactile data that can be used to train and improve downstream tactile policies.

\paragraph{Scaling Visuo-Tactile Data for Robot Learning.}
A key bottleneck  for tactile robot learning is the limited availability of synchronized visuo-tactile-action data. Real-robot tactile collection usually builds on standard robot-gripper-camera setups, such as DROID-style platforms~\cite{khazatsky2025droidlargescaleinthewildrobot}, with tactile sensors additionally mounted on the gripper fingers.
UMI-style and robot-free collection systems decouple demonstration collection from full robot teleoperation, and recent tactile extensions, including MimicTouch~\cite{yu2024mimictouch}, FreeTacMan~\cite{wu2025freetacman}, and exUMI~\cite{xu2025exumi}, among others~\cite{liu2025vitamin,xu2025dexumi,cheng2026tacumi}, enable visuo-tactile demonstrations with handheld or task-agnostic interfaces. 
These systems provide task-relevant trajectories, but scaling them still requires dedicated tactile hardware, careful synchronization, repeated physical contact, and may introduce embodiment or viewpoint gaps. 
Simulation offers another route: tactile simulators such as TacSL~\cite{akinola2025tacsl} and TacEx~\cite{nguyen2024tacex} enable scalable visuo-tactile sensor simulation and simulated contact-rich data collection~\cite{wang2022tacto,si2022taxim,si2024difftactile,li2025taccel,chen2026univtac}.
Yet simulated tactile data remain constrained by contact fidelity, asset realism, task diversity, and sim-to-real transfer. 
Recent robot world models~\cite{guo2026ctrlworld,guo2026vlaw} have explored using generated rollouts for policy evaluation or improvement, while large video and world foundation models provide strong generative priors for robot data synthesis.
However, most action-conditioned robot world models used for rollout generation remain primarily visual, which is insufficient for contact-rich manipulation where critical contact states are often weakly observable from cameras.
Visuo-tactile world-modeling work~\cite{higuera2026visuo} has shown that tactile observations can improve the physical consistency of imagined contact dynamics, but these models are not primarily designed as scalable data generators for downstream tactile policy learning. 
In contrast, ViTacWorld extends a large action-conditioned robot video prior into a visuo-tactile data generator, producing policy-improvement rollouts with aligned visual and tactile observations for contact-rich robot learning.

\section{Method}
\label{sec:ViTacWorld}

\begin{figure*}[ht]
    \centering
    \includegraphics[width=\textwidth]{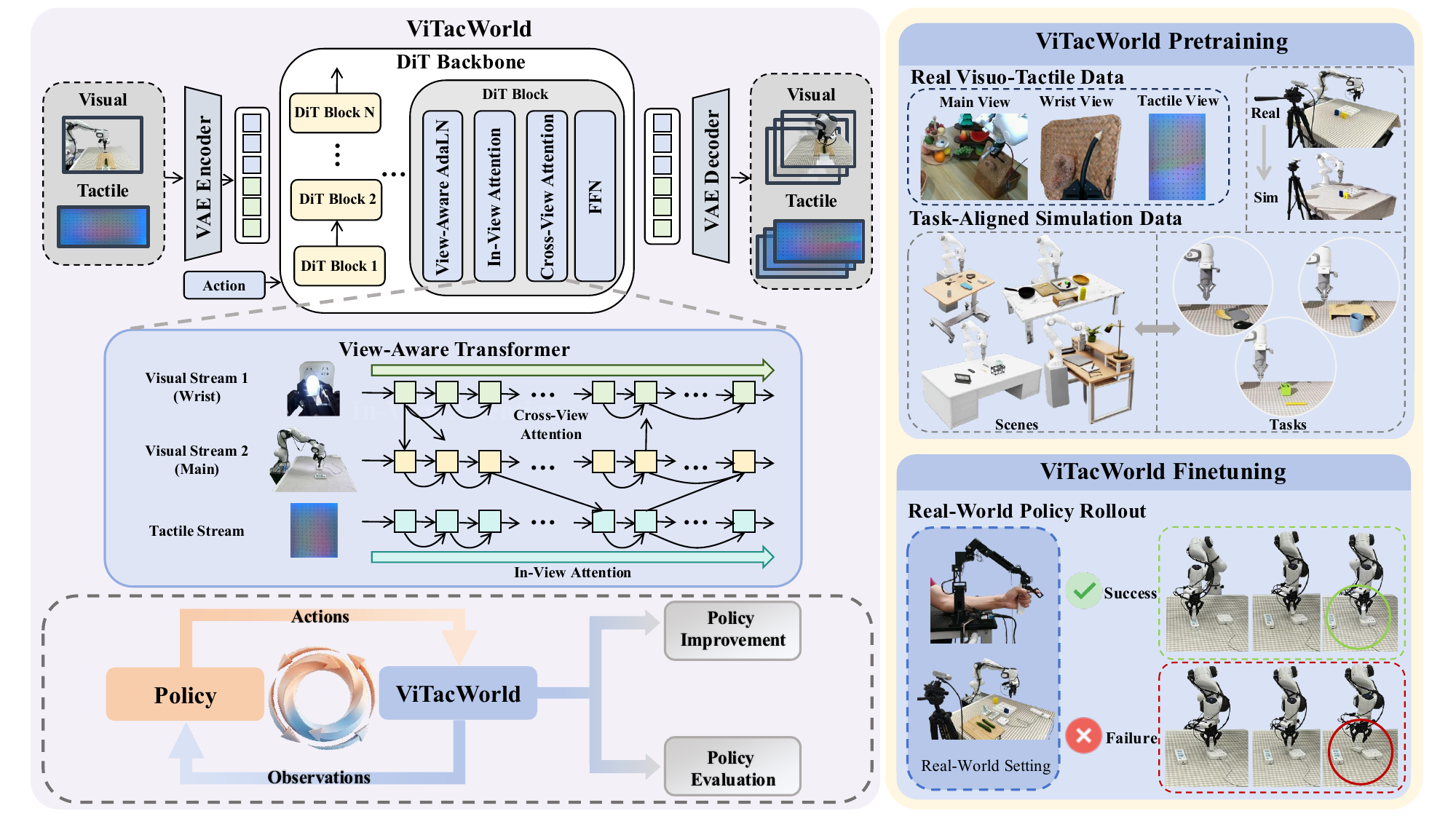}
\caption{
\textbf{Overview of ViTacWorld.}
Right side shows scaling training pipeline, where real visuo-tactile data and task-aligned simulation data are used for pretraining, followed by real robot task data and policy rollout data for target-domain tuning.
Left side illustrates the action-conditioned visuo-tactile world model and its rollout generation process. Visual and tactile streams are encoded into latent tokens and modeled with a view-aware DiT to produce policy-conditioned rollouts, which support both dataset augmentation and policy evaluation.
}
    \label{fig:pipeline}
\end{figure*}

Figure~\ref{fig:pipeline} summarizes the model architecture, the two-stage
training pipeline, and the two downstream uses of ViTacWorld.

\subsection{Problem Formulation}
\label{sec:problem_formulation}

We aim to learn an action-conditioned visuo-tactile world model for contact-rich manipulation. At time $t$, we represent the multimodal observation as $o_t=\{I_t^v\}_{v\in\mathcal{V}}$, where $\mathcal{V}=\{\mathrm{main},\mathrm{wrist},\mathrm{tactile}\}$ denotes the external camera, wrist camera, and image-like tactile stream, respectively. Given an $H$-step world-model action sequence $u_{t:t+H-1}$, where each action is represented as relative end-effector motion with a gripper command, the model predicts future visual and tactile observations,
\begin{equation}
\hat{o}_{t+1:t+H}=f_\theta(o_t,u_{t:t+H-1},m),
\end{equation}
where $m\in\{0,1\}^{|\mathcal{V}|}$ is a view-presence mask for heterogeneous training data. 

\subsection{Action-Conditioned Visuo-Tactile World Model}
\label{sec:action_conditioned_vitac_world_model}

ViTacWorld extends a pretrained action-conditioned robot video world model to visuo-tactile generation while preserving its backbone and action-conditioning pathway. For each training sample, the main camera, wrist camera, and tactile observations are separately encoded by the VAE~\cite{kingma2013auto} encoder into latent tokens and assembled as a unified visuo-tactile sequence. We then adapt the action-conditioned DiT~\cite{peebles2023scalable} with stream-aware modulation and explicit cross-stream exchange, allowing the model to reuse robot motion priors while generating aligned visual and tactile observations.

To make the DiT aware of the source stream of each latent token, we introduce stream identity embeddings $e^v$ for the visual and tactile streams. These embeddings are projected into the same AdaLN~\cite{perez2018film} modulation path as timestep and action conditioning. For each block $b$, stream-aware self-attention is first applied within each stream:
\begin{equation}
\tilde{Z}_b^v=
\mathrm{SelfAttn}_v
\left(
\mathrm{AdaLN}(Z_{b-1}^v; c_b + P(e^v))
\right),
\quad v\in\mathcal{V},
\end{equation}
where $Z_b^v$ denotes the latent tokens of stream $v$, $c_b$ denotes the timestep-action conditioning, and $P$ is a zero-initialized projection. Then an explicit cross-view attention module exchanges information across streams:
\begin{equation}
Z_b^v=
\mathrm{CrossViewAttn}_v
\left(
\tilde{Z}_b^v,\{\tilde{Z}_b^{v'}\}_{v'\neq v}
\right),
\quad v\in\mathcal{V}.
\end{equation}
This design avoids uncontrolled mixing between camera and tactile tokens during ordinary self-attention, while still allowing the tactile stream to exchange contact information with the visual streams. As a result, tactile is generated as a first-class observation stream that remains temporally aligned with the visual rollout.

\textbf{Action conditioning.} ViTacWorld follows the world-model action convention of the action-conditioned robot video prior. The action sequence is embedded through the inherited action-conditioning pathway and injected into the DiT through the original timestep/AdaLN~\cite{perez2018film} modulation path. To support multi-view visuo-tactile generation, we simply repeat the action conditioning across visual and tactile streams at the corresponding latent timestep.

\textbf{Training objective.}
ViTacWorld is trained with the latent denoising objective of the underlying video world model, applied to future visual and tactile latents. Let $z_0$ denote the target latent rollout and $z_\sigma$ its noisy version at diffusion noise level $\sigma$. The model predicts the denoising target conditioned on the current observation, robot actions, stream identities, and the view-presence mask:
\begin{equation}
\mathcal{L}_{\mathrm{wm}}
=
\mathrm{E}_{z_0,\sigma}
\left[
\left\|
D_\theta(z_\sigma,\sigma,o_t,u_{t:t+H-1},m)-z_0
\right\|_2^2
\right].
\end{equation}
The loss is applied only to future prediction frames and valid streams. This objective trains ViTacWorld to generate action-conditioned contact-rich rollouts with temporally aligned visual and tactile observations.
\subsection{ViTacWorld Pretraining and Fine-Tuning}
\label{sec:staged_adaptation}

A key design goal of ViTacWorld is to scale visuo-tactile world modeling beyond small task-specific tactile datasets. We therefore pretrain the model on large-scale public and simulated visuo-tactile data, and then finetune it with real demonstrations and policy rollouts from our target setup.

\textbf{Pretraining with large-scale real and simulated data.}
In the first stage, we pretrain ViTacWorld with large-scale real and simulated visuo-tactile-action trajectories to learn broad action-conditioned visual-tactile dynamics before target-domain adaptation. To handle heterogeneous view coverage, we use a fixed stream layout with view-presence masks, allowing incomplete trajectories to contribute to pretraining without corrupting missing streams. 

We collect task-aligned simulated data as a complementary pretraining source to bridge large-scale public data and our target real-robot tasks. Our motivation is that, although full manipulation simulation still suffers from scene, dynamics, and embodiment level sim-to-real gaps, the visual appearance of image-like tactile observations can be rendered with relatively smaller modality gap when the sensor geometry and rendering pipeline are well aligned. This makes tactile simulation a useful source for enriching contact patterns during pretraining. The simulation is built in Isaac Sim, with tactile observations synthesized through the Xense tactile rendering pipeline. To align simulation with the real setup, we reconstruct target scenes and objects with 3D Gaussian scanning, calibrate the real camera extrinsics, and reproduce the camera-robot spatial layout in simulation. Beyond matching the real tasks, we also vary task assets and interaction instances under the same robot-sensor setup, providing additional object and task diversity while keeping the generated data close to the target deployment domain.

Stage-I pretraining mixes over 16K public real trajectories with over 5K
task-aligned simulated trajectories at an approximately $2{:}1$ sampling ratio.
We train on 13-frame windows sampled at 15~Hz for 30K steps with a global batch
size of 256 and learning rate $3\!\times\!10^{-5}$. Stage-II target-domain
fine-tuning uses a global batch size of 128, learning rate
$1\!\times\!10^{-5}$, and 7K steps.

\textbf{Real-world policy rollout and finetuning.}
In the second stage, we adapt ViTacWorld to the target real-world setup using expert demonstrations and policy rollouts collected on the real robot. Expert trajectories align the model with real task executions, while policy rollouts expose it to states induced by downstream policies, including both successful and failed contact-rich interactions. This tuning stage helps the world model better match the deployment distribution in which synthetic rollouts will later be generated.

Together, the two stages turn a visual robot world-model prior into a target-ready visuo-tactile generator with both broad visual-tactile generalization and real-domain adaptability.

\subsection{Visuo-Tactile Policy Improvement and Evaluation}
\label{sec:synthetic_data_generation}

After training, ViTacWorld is used as both a policy-conditioned generator and a lightweight evaluator of visuo-tactile manipulation. Given real initial observations and action chunks proposed by a downstream tactile policy, ViTacWorld autoregressively generates contact-rich visual-tactile rollouts. These imagined rollouts can be inspected for success or failure, and selected successful rollouts are converted into dream data to augment real demonstrations for policy training.

Let $\pi_\phi$ denote a downstream tactile policy that maps the current observation $o_t$ and task instruction $l$ to an $H$-step action chunk:
\begin{equation}
a_{t:t+H-1}\sim \pi_\phi(\cdot \mid o_t,l).
\end{equation}
When necessary, policy actions are converted to the world-model action space defined in Sec.~\ref{sec:problem_formulation}. Given the current observation and action sequence, ViTacWorld predicts an $H$-step future rollout:
\begin{equation}
\hat{o}_{t+1:t+H}=f_\theta(o_t,u_{t:t+H-1},m).
\end{equation}
We then use the generated observation $\hat{o}_{t+k}$ as the next policy input and repeat the process autoregressively. This receding-horizon loop produces policy-conditioned contact-rich rollouts that reflect states likely to be visited by the downstream policy. We filter generated rollouts according to task success and visual-tactile plausibility, convert selected rollouts into policy-training trajectories, and merge them with expert demonstrations:
\begin{equation}
\mathcal{D}_{\mathrm{aug}}
=
\mathcal{D}_{\mathrm{expert}}
\cup
\mathcal{D}_{\mathrm{dream}}.
\end{equation}
The augmented dataset can then be used to finetune downstream tactile policies. In addition, the same policy-in-the-loop rollouts can be inspected as imagined executions, allowing ViTacWorld to provide a lightweight evaluation signal before real-robot deployment.







\section{Experiments}
\label{sec:experiments}

Our experiments evaluate whether ViTacWorld-generated rollouts improve
contact-rich manipulation policies, whether imagined executions preserve
real-robot policy performance, and how training choices affect generation quality.

\subsection{Experimental Setup}
\label{sec:experimental_setup}

\textbf{Robot platform.}
All real-robot experiments are conducted on a Franka Panda robot equipped with a Robotiq 2F-85 parallel gripper. Xense optical tactile sensors are mounted on the gripper fingers to capture image-like tactile observations during contact-rich interaction. Visual observations are captured by an Intel RealSense D435 camera as the external third-person view and a ZED Mini camera as the wrist-mounted view. The tactile policies take synchronized third-person RGB, wrist RGB, tactile observations, robot proprioception, and task instructions as input. Expert demonstrations are collected through FACTR
teleoperation~\cite{liu2025factr}.

\textbf{Tasks.}
We evaluate on four contact-rich manipulation tasks: \textit{Charger Plugging}, \textit{Cucumber Peeling}, \textit{U-Block Insertion}, and \textit{Cuboid Insertion}. These tasks cover precise charger alignment and insertion, tool-object contact during peeling, and object-slot insertion with tight geometric constraints. Among them, \textit{Charger Plugging} is particularly challenging because small pose errors can lead to failed insertion even when the visual trajectory appears similar.

\textbf{Expert and rollout data.}
We collect 300 real expert demonstrations for policy warm-up, including 120 charger plugging, 50 cucumber peeling, 62 U-block insertion, and 68 cuboid insertion demonstrations. After warming up a tactile $\pi_{0.5}$ policy~\cite{intelligence2025pi}, we further collect 50 real policy rollouts per task, including both successful and failed executions. These rollouts are used to tune ViTacWorld to the state distribution induced by the downstream policy.

\textbf{Visuo-Tactile data generation.}
After ViTacWorld is trained, we use the warmed-up tactile $\pi_{0.5}$ policy as the rollout policy to generate candidate visuo-tactile trajectories. We select 200 successful rollouts across the four tasks and merge them with the expert demonstrations. 
Candidate rollouts are retained only when a human annotator verifies fixed
task-completion criteria and visual--tactile plausibility; task-specific rules
include complete insertion and continuous valid tool--object contact.

\textbf{Downstream policies.}
We evaluate a vision-only $\pi_{0.5}$ baseline, a tactile $\pi_{0.5}$ policy,
and ACT + tactile~\cite{zhao2023learning}. All three are trained at the
expert-only, round-1, and round-2 stages. The vision-only baseline uses the
visual observations and actions from the selected rollouts without tactile
input, whereas the tactile policies also use the generated tactile stream.
Round~2 adds another 200 selected rollouts generated by the round-1 tactile
$\pi_{0.5}$ policy, retaining the 200 round-1 rollouts and 300 expert
demonstrations for a total of 700 training trajectories.
Thus, Table~\ref{tab:real_robot_success} compares expert-only training with
visual-only and visuo-tactile rollout augmentation. The tactile $\pi_{0.5}$
policy generates the rollouts used to finetune all downstream policies.

\textbf{Evaluation protocol.}
We report real-robot success rates over 50 trials for each policy, training
stage, and task, using the same task setup and initial-state distribution.
A trial is successful only when the robot completes the task-specific goal,
such as full insertion, sustained peeling contact, or placement into the
target slot. The average is the unweighted mean across the four tasks.

\subsection{Policy Improvement with Generated Visuo-Tactile Data}
\label{sec:policy_improvement_results}

\begin{table*}[t]\vspace*{6pt} 
\centering
\caption{Real-robot success rates (\%) over 50 trials per task, policy, and
training stage. Avg. is the four-task mean; bold marks the best result per column.}
\label{tab:real_robot_success}
\small
\setlength{\tabcolsep}{4pt}
\begin{tabular*}{\textwidth}{l@{\extracolsep{\fill}}lccccc}
\toprule
Data source & Method 
& \shortstack{Charger\\Plugging} 
& \shortstack{Cucumber\\Peeling} 
& \shortstack{U-Block\\Insertion} 
& \shortstack{Cuboid\\Insertion} 
& Avg. \\
\midrule

\multirow{3}{*}{\textit{Expert only}} 
& ACT + tactile & 4 & 8 & 32 & 24 & 17.0 \\
& $\pi_{0.5}$ & 18 & 28 & 48 & 28 & 30.5 \\
& $\pi_{0.5}$ + tactile & 20 & 44 & 72 & 36 & 43.0 \\

\midrule

\multirow{3}{*}{\shortstack[l]{\textit{Expert +}\\\textit{Round-1 rollouts}}}
& ACT + tactile & 14 & 22 & 36 & 32 & 26.0 \\
& $\pi_{0.5}$ & 34 & 60 & 68 & 42 & 51.0 \\
& $\pi_{0.5}$ + tactile & 44 & 82 & 80 & 64 & 67.5 \\
\midrule
\multirow{3}{*}{\shortstack[l]{\textit{Expert +}\\\textit{Round-2 rollouts}}}
& ACT + tactile & 26 & 44 & 68 & 50 & 47.0 \\
& $\pi_{0.5}$ & 44 & 72 & 82 & 62 & 65.0 \\
& $\pi_{0.5}$ + tactile & \textbf{64} & \textbf{86} & \textbf{94} & \textbf{74} & \textbf{79.5} \\
\bottomrule
\end{tabular*}

\par\vspace*{-6pt}\end{table*} 

ViTacWorld-generated rollouts improve tactile-policy performance across all
four tasks (Table~\ref{tab:real_robot_success}). The mean success rate of
tactile $\pi_{0.5}$ rises from 43.0\% with expert demonstrations to 67.5\%
and 79.5\% after successive augmentation rounds. ACT + tactile also benefits,
improving from 17.0\% to 47.0\%. These results demonstrate that generated
visuo-tactile trajectories provide useful training supervision across two
policy architectures, with further gains from iterative augmentation.

To examine the role of generated tactile observations, we compare the
vision-only and tactile $\pi_{0.5}$ policies trained on the same selected
rollouts, with the tactile stream omitted for the vision-only baseline.
With the added rollout count and visual--action sequences matched, the
tactile policy achieves a larger four-task gain over its own expert-only
baseline: 24.5 versus 20.5 percentage points after round~1, and 36.5 versus
34.5 after round~2. The tactile policy also maintains higher success on
every task at every training stage. These comparisons support the utility
of retaining generated tactile observations for policy augmentation,
alongside the benefit of additional visual--action training data.

\begin{figure}[t]\vspace*{6pt} 
    \centering
    \includegraphics[
        width=\linewidth,
    ]{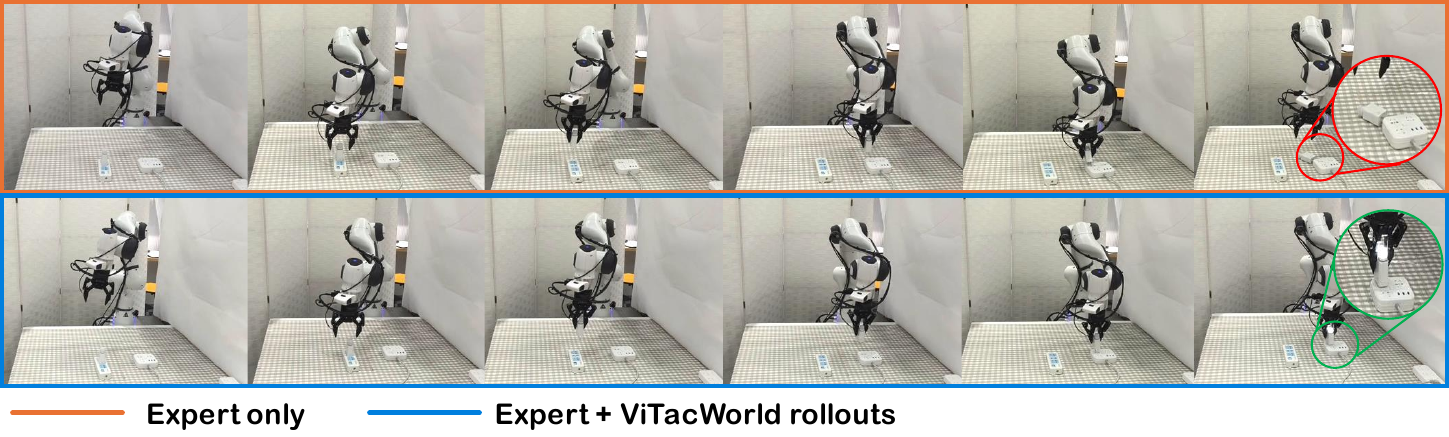}
    \caption{Qualitative comparison of real-robot rollouts using the same $\pi_{0.5}$ + tactile policy architecture. The top row shows the policy trained only with expert demonstrations, while the bottom row shows the policy trained with the same expert demonstrations augmented by ViTacWorld-generated rollouts.}
    \label{fig:policy_aug_comparison}\par\vspace*{-6pt}
\end{figure}

Figure~\ref{fig:policy_aug_comparison} illustrates the corresponding changes
in real-robot behavior. The expert-only policy can fail through inaccurate
grasping or poor alignment before insertion. After augmentation, the tactile
policy shows more stable grasping and contact-phase corrections, including
small re-alignment motions after an initially misaligned insertion attempt.
These examples complement the success-rate results by showing how additional
visuo-tactile rollouts can improve both approach behavior and contact adjustment.

\begin{figure}[t]
  \centering
  \includegraphics[width=\linewidth]{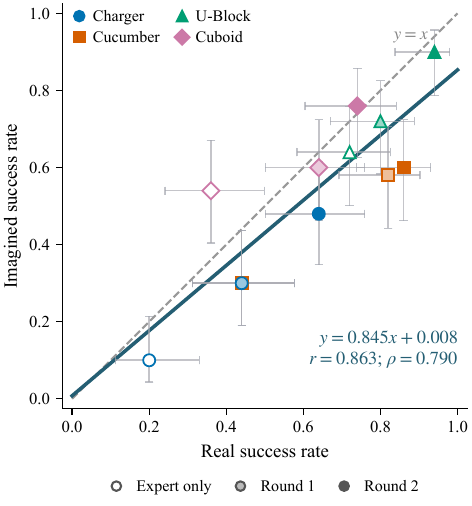}
  \caption{Real-versus-imagined success rates for three tactile $\pi_{0.5}$
  checkpoints on four tasks (12 points), each based on 50 matched initial
  states. Colors and marker shapes denote tasks; open, light-filled, and
  solid markers denote the expert-only, round-1, and round-2 stages.
  Bars are marginal 95\% Wilson confidence intervals for each success rate.
  The dashed line is $y=x$ and the solid line is the least-squares fit.}
  \label{fig:policy_eval_correlation}
\end{figure}

\subsection{Policy Evaluation with Imagined Rollouts}
\label{sec:policy_evaluator}
We evaluate the expert-only, round-1, and round-2 tactile $\pi_{0.5}$
checkpoints on all four tasks, reusing the corresponding real-robot trials
reported in Table~\ref{tab:real_robot_success}. For each checkpoint--task
combination, we use the recorded initial state and observation from each of
the 50 trials to initialize one imagined execution with the same policy
checkpoint. This gives 50 real--imagined pairs per combination and
600 pairs in total. A human annotator applies the same fixed task-completion
criteria to real and imagined executions; each success rate is the fraction
of successful executions among the 50 trials.

Figure~\ref{fig:policy_eval_correlation} compares the resulting 12 pairs of
aggregate success rates. Real and imagined performance are positively
associated (Pearson $r=0.863$; Spearman $\rho=0.790$), with a mean absolute
error of 12.33 percentage points. Within every task, both evaluations rank
the checkpoints in the order expert-only $<$ round~1 $<$ round~2. However,
ViTacWorld underestimates success by 9.0 points on average, with a 26-point
error for round-2 Cucumber Peeling. These results support relative comparison
among the evaluated checkpoints, but do not establish exact success-rate
calibration or trial-level outcome agreement from the aggregate rates alone.

\begin{figure}[t]
    \centering
    \includegraphics[width=\linewidth]{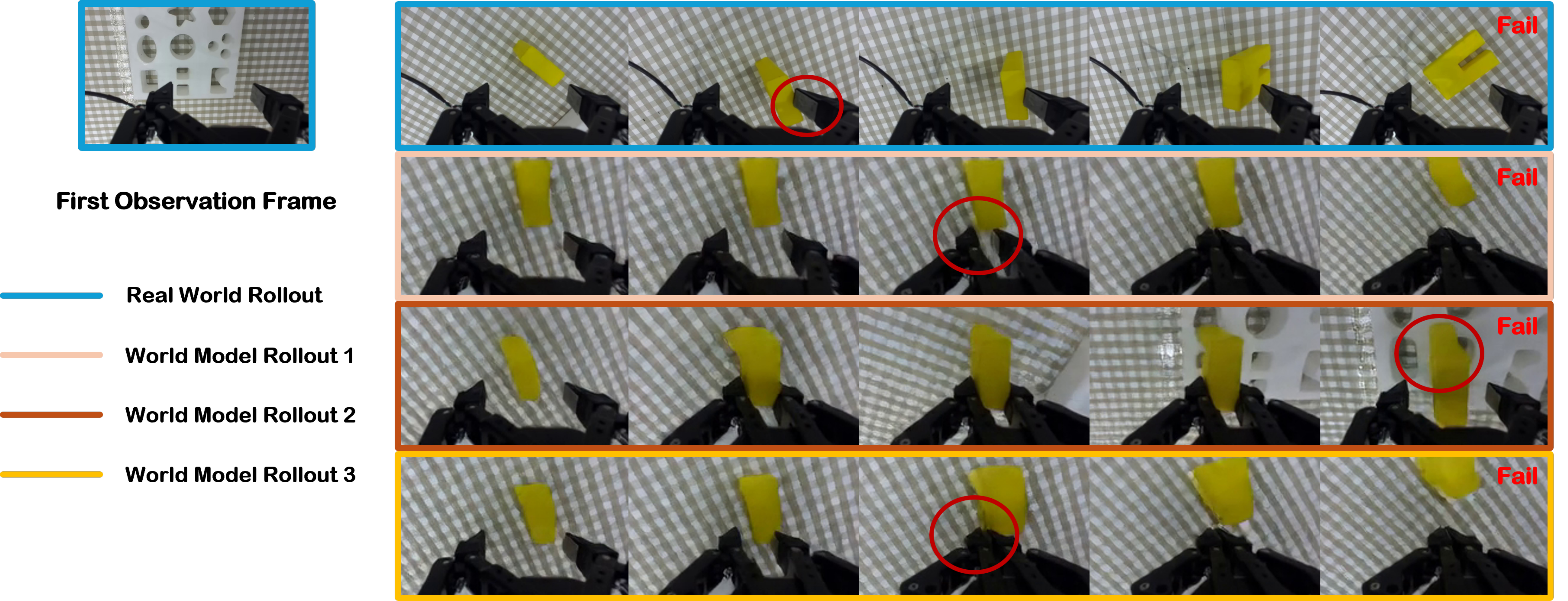}
    \vspace{-0.8em}
    \caption{Matched qualitative policy-evaluation example on U-Block
    insertion. The shared initial observation is shown at left. The top row is
    the failed real-robot execution; the following three rows are independent
    ViTacWorld rollouts, all of which also predict failure. This example
    illustrates outcome agreement and is separate from the 50-trial
    quantitative evaluation.}
    \label{fig:wm_eval_ublock_failure}
\end{figure}

Separately, Figure~\ref{fig:wm_eval_ublock_failure} illustrates a matched
qualitative U-Block failure example. From the same initial observation, the real execution and all three
independently sampled imagined executions fail the task. The generated
trajectories need not be frame-identical to the real rollout; the relevant
signal is agreement on the task-level outcome under the matched policy and
initial condition.

\begin{figure*}[t]
    \centering
    \vspace*{6pt}\includegraphics[width=\textwidth]{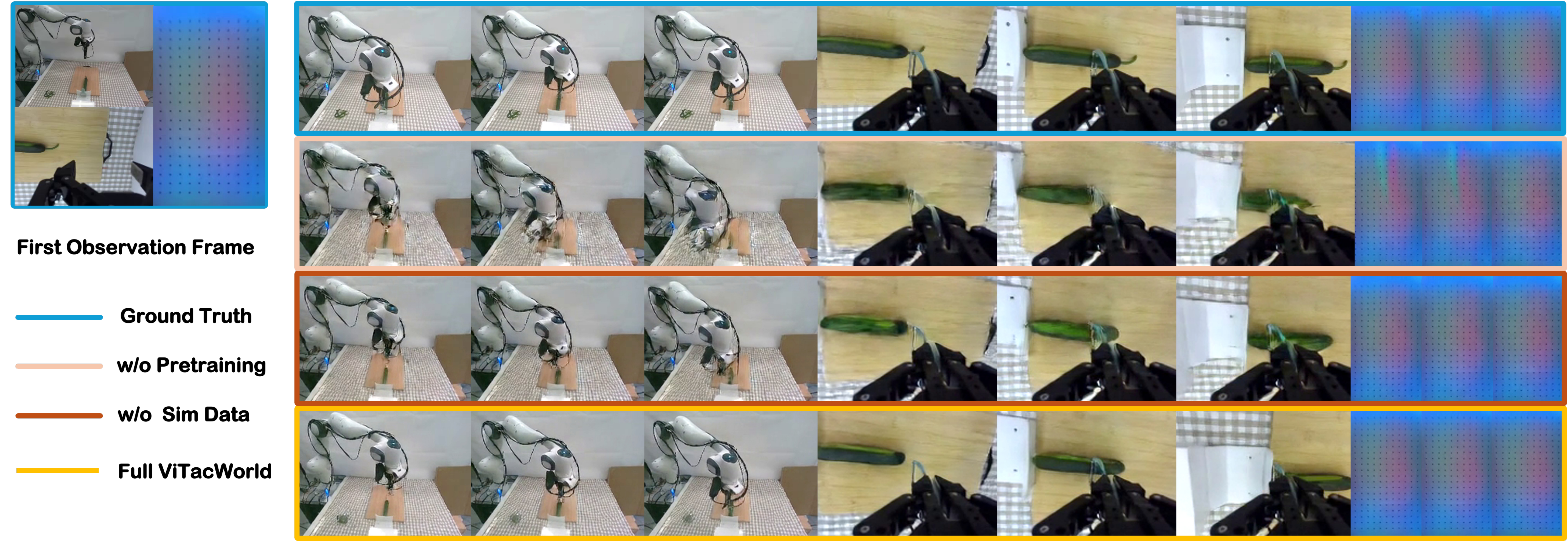}
    \caption{Qualitative world-model predictions on held-out validation clips. The model receives the first observation frame and the logged action sequence, then predicts future visual and tactile observations.}
    \label{fig:wm_validation_rollouts}\par\vspace*{-6pt}
\end{figure*}

\subsection{Visuo-Tactile World Model Quality Analysis}
\label{sec:wm_quality_analysis}
To reduce score inflation from prolonged non-contact segments, we evaluate
all variants on the same held-out real-world clips spanning contact
transitions, rather than full trajectories. Given the first observation
and logged action sequence, each model predicts future observations; we
compute view-wise PSNR, SSIM, and LPIPS against the corresponding
ground-truth future frames. These image-based measures complement the
qualitative predictions and downstream policy evaluation.

As shown in Table~\ref{tab:wm_quality}, pretraining substantially improves generation quality over direct real-data tuning, indicating that large-scale visuo-tactile data helps the world model learn more stable action-conditioned dynamics beyond the limited real tuning set. Adding task-aligned simulated tactile trajectories further improves the model over real-data pretraining alone, suggesting that simulation aligned with our real scenes and tasks helps adapt the world model to the target robot domain.

Replacing cross-view attention with self-attention throughout the model increases LPIPS from 0.0513 to 0.0611 in the main view, from 0.0725 to 0.0854 in the wrist view, and from 0.0157 to 0.0163 in the tactile stream. Main-view SSIM is nearly unchanged (0.8286 versus 0.8290). These results support the contribution of explicit cross-view attention to perceptual prediction quality across the three streams.

\begin{table}[t]
\centering
\caption{World-model prediction quality on held-out contact-transition clips.
Bold marks the best result per view and metric.}
\label{tab:wm_quality}
\small
\setlength{\tabcolsep}{3pt}
\begin{tabular*}{\linewidth}{@{}l@{\extracolsep{\fill}}lccc@{}}
\toprule
View & Model variant & PSNR $\uparrow$ & SSIM $\uparrow$ & LPIPS $\downarrow$ \\
\midrule
\multirow{4}{*}{Main}
& w/o pretraining      & 22.718 & 0.7859 & 0.0781 \\
& w/o task-aligned sim & 23.128 & 0.8011 & 0.0687 \\
& w/o cross-view attn. & 23.653 & \textbf{0.8290} & 0.0611 \\
& Full ViTacWorld     & \textbf{24.258} & 0.8286 & \textbf{0.0513} \\
\midrule
\multirow{4}{*}{Wrist}
& w/o pretraining      & 21.080 & 0.6649 & 0.1084 \\
& w/o task-aligned sim & 21.434 & 0.6869 & 0.0901 \\
& w/o cross-view attn. & 21.853 & 0.6913 & 0.0854 \\
& Full ViTacWorld     & \textbf{21.925} & \textbf{0.6962} & \textbf{0.0725} \\
\midrule
\multirow{4}{*}{Tactile}
& w/o pretraining      & 34.967 & 0.9204 & 0.0211 \\
& w/o task-aligned sim & 35.127 & 0.9296 & 0.0179 \\
& w/o cross-view attn. & 35.220 & 0.9310 & 0.0163 \\
& Full ViTacWorld     & \textbf{35.225} & \textbf{0.9318} & \textbf{0.0157} \\
\bottomrule
\end{tabular*}
\end{table}

We also provide qualitative visualizations in Figure~\ref{fig:wm_validation_rollouts}. The held-out validation examples show that pretraining and task-aligned simulation improve generation quality across visual and tactile views. Compared with ablated variants, the full model produces clearer rollouts with better preserved object geometry, more consistent end-effector motion, and fewer hallucinated contact states.


\FloatBarrier\section{Conclusion}
\label{sec:conclusion}

In this paper, we present ViTacWorld, an action-conditioned visuo-tactile world model for contact-rich robot manipulation. ViTacWorld generates temporally aligned visual and tactile rollouts under robot actions, serving both as a scalable source of synthetic visuo-tactile-action data and as a lightweight policy evaluation tool. We train the model with large-scale real and simulated visuo-tactile trajectories, followed by real-world policy-rollout fine-tuning. Experiments on contact-rich tasks show that ViTacWorld produces policy-improvement rollouts and supports imagination-based policy evaluation.




\section*{Acknowledgments}
This work was supported by the National Natural Science Foundation of China (62406195), the HPC Platform of ShanghaiTech University, the Key Laboratory of Intelligent Perception and Human-Machine Collaboration (Shanghaitech University), Ministry of Education, and the Shanghai Engineering Research Center of Intelligent Vision and Imaging. This work was also supported in part by computational resources provided by Fcloud CO., LTD.

\bibliographystyle{IEEEtran}
\bibliography{references}

\begin{thebibliography}{10}
\providecommand{\url}[1]{#1}
\csname url@rmstyle\endcsname
\providecommand{\newblock}{\relax}
\providecommand{\bibinfo}[2]{#2}
\providecommand\BIBentrySTDinterwordspacing{\spaceskip=0pt\relax}
\providecommand\BIBentryALTinterwordstretchfactor{4}
\providecommand\BIBentryALTinterwordspacing{\spaceskip=\fontdimen2\font plus
\BIBentryALTinterwordstretchfactor\fontdimen3\font minus
  \fontdimen4\font\relax}
\providecommand\BIBforeignlanguage[2]{{%
\expandafter\ifx\csname l@#1\endcsname\relax
\typeout{** WARNING: IEEEtran.bst: No hyphenation pattern has been}%
\typeout{** loaded for the language `#1'. Using the pattern for}%
\typeout{** the default language instead.}%
\else
\language=\csname l@#1\endcsname
\fi
#2}}

\bibitem{dong2017improvedgelsight}
S.~Dong, W.~Yuan, and E.~H. Adelson, ``Improved gelsight tactile sensor for
  measuring geometry and slip,'' in \emph{2017 IEEE/RSJ International
  Conference on Intelligent Robots and Systems (IROS)}, 2017, pp. 137--144.

\bibitem{yuan2017gelsight}
W.~Yuan, S.~Dong, and E.~H. Adelson, ``Gelsight: High-resolution tactile
  sensors for perceiving physical properties,'' \emph{IEEE Robotics \&
  Automation Magazine}, vol.~24, no.~3, pp. 66--77, 2017.

\bibitem{akinola2025tacsl}
I.~Akinola, J.~Xu, J.~Carius, D.~Fox, and Y.~Narang, ``Tacsl: A library for
  visuotactile sensor simulation and learning,'' \emph{IEEE Transactions on
  Robotics}, 2025.

\bibitem{nguyen2024tacex}
D.~H. Nguyen, T.~Schneider, G.~Duret, A.~Kshirsagar, B.~Belousov, and
  J.~Peters, ``Tacex: Gelsight tactile simulation in isaac sim--combining
  soft-body and visuotactile simulators,'' \emph{arXiv preprint
  arXiv:2411.04776}, 2024.

\bibitem{chen2026univtac}
B.~Chen, W.~Wan, T.~Chen, X.~Guo, C.~Xu, Y.~Qi, H.~Zhang, L.~Wu, T.~Xu, Z.~Li,
  \emph{et~al.}, ``Univtac: A unified simulation platform for visuo-tactile
  manipulation data generation, learning, and benchmarking,'' \emph{arXiv
  preprint arXiv:2602.10093}, 2026.

\bibitem{chi2024umi}
C.~Chi, Z.~Xu, C.~Pan, E.~Cousineau, B.~Burchfiel, S.~Feng, R.~Tedrake, and
  S.~Song, ``Universal manipulation interface: In-the-wild robot teaching
  without in-the-wild robots,'' in \emph{Robotics: Science and Systems (RSS)},
  2024.

\bibitem{zhaxizhuoma2025fastumi}
Z.~Zhaxizhuoma, K.~Liu, C.~Guan, Z.~Jia, Z.~Wu, X.~Liu, T.~Wang, S.~Liang,
  \emph{et~al.}, ``Fast-umi: A scalable and hardware-independent universal
  manipulation interface with dataset,'' in \emph{Conference on Robot Learning
  (CoRL)}, 2025.

\bibitem{wu2025freetacman}
L.~Wu, C.~Yu, J.~Ren, L.~Chen, Y.~Jiang, R.~Huang, G.~Gu, and H.~Li,
  ``Freetacman: Robot-free visuo-tactile data collection system for
  contact-rich manipulation,'' \emph{arXiv preprint arXiv:2506.01941}, 2025.

\bibitem{xu2025exumi}
Y.~Xu, L.~Wei, P.~An, Q.~Zhang, and Y.-L. Li, ``exumi: Extensible robot
  teaching system with action-aware task-agnostic tactile representation,''
  \emph{arXiv preprint arXiv:2509.14688}, 2025.

\bibitem{blattmann2023stablevideo}
A.~Blattmann, T.~Dockhorn, S.~Kulal, D.~Mendelevitch, M.~Kilian, D.~Lorenz,
  Y.~Levi, Z.~English, V.~Voleti, A.~Letts, \emph{et~al.}, ``Stable video
  diffusion: Scaling latent video diffusion models to large datasets,''
  \emph{arXiv preprint arXiv:2311.15127}, 2023.

\bibitem{chi2025wow}
X.~Chi, P.~Jia, C.-K. Fan, X.~Ju, W.~Mi, K.~Zhang, Z.~Qin, W.~Tian, K.~Ge,
  H.~Li, \emph{et~al.}, ``Wow: Towards a world omniscient world model through
  embodied interaction,'' \emph{arXiv preprint arXiv:2509.22642}, 2025.

\bibitem{wan2025wan}
T.~Wan, A.~Wang, B.~Ai, B.~Wen, C.~Mao, C.-W. Xie, D.~Chen, F.~Yu, H.~Zhao,
  J.~Yang, \emph{et~al.}, ``Wan: Open and advanced large-scale video generative
  models,'' \emph{arXiv preprint arXiv:2503.20314}, 2025.

\bibitem{ali2025cosmospredict25}
A.~Ali, J.~Bai, M.~Bala, Y.~Balaji, A.~Blakeman, T.~Cai, J.~Cao, T.~Cao,
  E.~Cha, Y.-W. Chao, \emph{et~al.}, ``World simulation with video foundation
  models for physical ai,'' \emph{arXiv preprint arXiv:2511.00062}, 2025.

\bibitem{huang2025particleformer}
S.~Huang, Q.~Chen, X.~Zhang, J.~Sun, and M.~Schwager, ``Particleformer: A 3d
  point cloud world model for multi-object, multi-material robotic
  manipulation,'' \emph{arXiv preprint arXiv:2506.23126}, 2025.

\bibitem{higuera2026visuo}
C.~Higuera, S.~Arnaud, B.~Boots, M.~Mukadam, F.~R. Hogan, and F.~Meier,
  ``Visuo-tactile world models,'' \emph{arXiv preprint arXiv:2602.06001}, 2026.

\bibitem{zheng2026omnivta}
Y.~Zheng, S.~Gu, W.~Li, Y.~Zheng, Y.~Zang, S.~Tian, X.~Li, C.~Hao, C.~Gao,
  S.~Liu, \emph{et~al.}, ``Omnivta: Visuo-tactile world modeling for
  contact-rich robotic manipulation,'' \emph{arXiv preprint arXiv:2603.19201},
  2026.

\bibitem{yuan2026vtam}
H.~Yuan, W.~Yi, Z.~Zhang, W.~Chen, Y.~Mo, J.~Yin, X.~Li, X.~Zeng, C.~Wen,
  C.~Lu, \emph{et~al.}, ``Vtam: Video-tactile-action models for complex
  physical interaction beyond vlas,'' \emph{arXiv preprint arXiv:2603.23481},
  2026.

\bibitem{chi2025diffusion}
C.~Chi, Z.~Xu, S.~Feng, E.~Cousineau, Y.~Du, B.~Burchfiel, R.~Tedrake, and
  S.~Song, ``Diffusion policy: Visuomotor policy learning via action
  diffusion,'' \emph{The International Journal of Robotics Research}, vol.~44,
  no. 10-11, pp. 1684--1704, 2025.

\bibitem{ze20243d}
Y.~Ze, G.~Zhang, K.~Zhang, C.~Hu, M.~Wang, and H.~Xu, ``3d diffusion policy:
  Generalizable visuomotor policy learning via simple 3d representations,''
  \emph{arXiv preprint arXiv:2403.03954}, 2024.

\bibitem{wu2025afforddp}
S.~Wu, Y.~Zhu, Y.~Huang, K.~Zhu, J.~Gu, J.~Yu, Y.~Shi, and J.~Wang, ``Afforddp:
  Generalizable diffusion policy with transferable affordance,'' in
  \emph{Proceedings of the Computer Vision and Pattern Recognition Conference},
  2025, pp. 6971--6980.

\bibitem{Huang20243DViTacLF}
B.~Huang, Y.~Wang, X.~Yang, Y.~Luo, and Y.~Li, ``3d-vitac: Learning
  fine-grained manipulation with visuo-tactile sensing,'' in \emph{Conference
  on Robot Learning}, 2024.

\bibitem{xue2025reactive}
H.~Xue, J.~Ren, W.~Chen, G.~Zhang, Y.~Fang, G.~Gu, H.~Xu, and C.~Lu, ``Reactive
  diffusion policy: Slow-fast visual-tactile policy learning for contact-rich
  manipulation,'' in \emph{Proceedings of Robotics: Science and Systems (RSS)},
  2025.

\bibitem{zhao2025polytouch}
J.~Zhao, N.~Kuppuswamy, S.~Feng, B.~Burchfiel, and E.~Adelson, ``Polytouch: A
  robust multi-modal tactile sensor for contact-rich manipulation using
  tactile-diffusion policies,'' in \emph{2025 IEEE International Conference on
  Robotics and Automation (ICRA)}.\hskip 1em plus 0.5em minus 0.4em\relax IEEE,
  2025, pp. 104--110.

\bibitem{bi2025vlatouch}
J.~Bi, K.~Y. Ma, C.~Hao, M.~S. Zheng, and H.~Soh, ``Vla-touch: Enhancing
  vision-language-action model with dual-level tactile feedback,'' \emph{IEEE
  Robotics and Automation Letters}, 2026.

\bibitem{huang2025tactilevla}
J.~Huang, S.~Wang, F.~Lin, Y.~Hu, C.~Wen, and Y.~Gao, ``Tactile-vla: unlocking
  vision-language-action model's physical knowledge for tactile
  generalization,'' \emph{arXiv preprint arXiv:2507.09160}, 2025.

\bibitem{yu2025forcevla}
J.~Yu, H.~Liu, Q.~Yu, J.~Ren, C.~Hao, H.~Ding, G.~Huang, G.~Huang, Y.~Song,
  P.~Cai, \emph{et~al.}, ``Forcevla: Enhancing vla models with a force-aware
  moe for contact-rich manipulation,'' \emph{Advances in Neural Information
  Processing Systems}, vol.~38, pp. 93\,409--93\,439, 2026.

\bibitem{li2026forcevla2}
Y.~Li, H.~Jiang, J.~Xia, H.~Zhang, J.~Du, Y.~Zhou, J.~Zeng, C.~Hao, J.~Ren,
  Q.~Yu, \emph{et~al.}, ``Forcevla2: Unleashing hybrid force-position control
  with force awareness for contact-rich manipulation,'' \emph{arXiv preprint
  arXiv:2603.15169}, 2026.

\bibitem{hao2025tla}
P.~Hao, C.~Zhang, D.~Li, X.~Cao, X.~Hao, S.~Cui, and S.~Wang, ``Tla:
  Tactile-language-action model for contact-rich manipulation,'' \emph{arXiv
  preprint arXiv:2503.08548}, 2025.

\bibitem{morissette2026tactile}
C.~Morissette, A.~Abyaneh, W.-D. Chang, A.~Houssaini, D.~Meger, H.-C. Lin,
  J.~Tremblay, and G.~Dudek, ``Tactile modality fusion for
  vision-language-action models,'' \emph{arXiv preprint arXiv:2603.14604},
  2026.

\bibitem{zhang2025vtla}
C.~Zhang, P.~Hao, X.~Cao, X.~Hao, S.~Cui, and S.~Wang, ``Vtla:
  Vision-tactile-language-action model with preference learning for insertion
  manipulation,'' \emph{Biomimetic Intelligence and Robotics}, p. 100333, 2026.

\bibitem{ye2026dreamtacvla}
G.~Ye, Z.~Zhang, X.~Zhao, S.~Wu, H.~Lu, S.~Lu, and H.~Liu, ``Learning to feel
  the future: Dreamtacvla for contact-rich manipulation,'' \emph{arXiv preprint
  arXiv:2512.23864}, 2025.

\bibitem{tian2026vtwam}
S.~Tian, Y.~Zheng, Y.~Zheng, S.~Gu, Y.~Zang, Y.~Qin, W.~Li, H.~Li, W.~Ding, and
  D.~Zhao, ``{VT-WAM}: Visual-tactile world action model for contact-rich
  manipulation,'' \emph{arXiv preprint arXiv:2607.02503}, 2026.

\bibitem{zang2026tacforesight}
Y.~Zang, Y.~Zheng, X.~Nie, Y.~Zheng, S.~Tian, S.~Gu, C.~Gao, Z.~Wang, S.~Yan,
  and W.~Ding, ``{TacForeSight}: Force-guided tactile world model for
  contact-rich manipulation,'' \emph{arXiv preprint arXiv:2606.11184}, 2026.

\bibitem{zhou2026touchworld}
J.~Zhou, F.~Hong, Y.~Li, Y.~Zhao, Y.~Cen, Z.~Liu, J.~Huang, Z.~Chen, R.~Zhang,
  W.~Zhu, X.~Song, and S.~Yang, ``{TouchWorld}: A predictive and reactive
  tactile foundation model for dexterous manipulation,'' \emph{arXiv preprint
  arXiv:2607.07287}, 2026.

\bibitem{khazatsky2025droidlargescaleinthewildrobot}
A.~Khazatsky, K.~Pertsch, S.~Nair, A.~Balakrishna, S.~Dasari, S.~Karamcheti,
  S.~Nasiriany, M.~K. Srirama, L.~Y. Chen, K.~Ellis, \emph{et~al.}, ``Droid: A
  large-scale in-the-wild robot manipulation dataset,'' in \emph{Robotics:
  Science and Systems}, 2024.

\bibitem{yu2024mimictouch}
K.~Yu, Y.~Han, Q.~Wang, V.~Saxena, D.~Xu, and Y.~Zhao, ``Mimictouch: Leveraging
  multi-modal human tactile demonstrations for contact-rich manipulation,'' in
  \emph{Conference on Robot Learning}.\hskip 1em plus 0.5em minus 0.4em\relax
  PMLR, 2025, pp. 4844--4865.

\bibitem{liu2025vitamin}
F.~Liu, C.~Li, Y.~Qin, J.~Xu, P.~Abbeel, and R.~Chen, ``Vitamin: Learning
  contact-rich tasks through robot-free visuo-tactile manipulation interface,''
  \emph{arXiv preprint arXiv:2504.06156}, 2025.

\bibitem{xu2025dexumi}
M.~Xu, H.~Zhang, Y.~Hou, Z.~Xu, L.~Fan, M.~Veloso, and S.~Song, ``Dexumi: Using
  human hand as the universal manipulation interface for dexterous
  manipulation,'' in \emph{Conference on Robot Learning}.\hskip 1em plus 0.5em
  minus 0.4em\relax PMLR, 2025, pp. 437--459.

\bibitem{cheng2026tacumi}
T.~Cheng, K.~Chen, L.~Chen, L.~Zhang, Y.~Zhang, Y.~Ling, M.~Hamad, Z.~Bing,
  F.~Wu, K.~Sharma, \emph{et~al.}, ``Tacumi: A multi-modal universal
  manipulation interface for contact-rich tasks,'' \emph{arXiv preprint
  arXiv:2601.14550}, 2026.

\bibitem{wang2022tacto}
S.~Wang, M.~Lambeta, P.-W. Chou, and R.~Calandra, ``Tacto: A fast, flexible,
  and open-source simulator for high-resolution vision-based tactile sensors,''
  \emph{IEEE Robotics and Automation Letters}, vol.~7, no.~2, pp. 3930--3937,
  2022.

\bibitem{si2022taxim}
Z.~Si and W.~Yuan, ``Taxim: An example-based simulation model for gelsight
  tactile sensors,'' \emph{IEEE Robotics and Automation Letters}, vol.~7,
  no.~2, pp. 2361--2368, 2022.

\bibitem{si2024difftactile}
Z.~Si, G.~Zhang, Q.~Ben, B.~Romero, Z.~Xian, C.~Liu, and C.~Gan, ``Difftactile:
  A physics-based differentiable tactile simulator for contact-rich robotic
  manipulation,'' \emph{arXiv preprint arXiv:2403.08716}, 2024.

\bibitem{li2025taccel}
Y.~Li, W.~Du, C.~Yu, P.~Li, Z.~Zhao, T.~Liu, C.~Jiang, Y.~Zhu, and S.~Huang,
  ``Taccel: Scaling up vision-based tactile robotics via high-performance gpu
  simulation,'' \emph{Advances in Neural Information Processing Systems},
  vol.~38, pp. 94\,577--94\,604, 2026.

\bibitem{guo2026ctrlworld}
Y.~Guo, L.~X. Shi, J.~Chen, and C.~Finn, ``Ctrl-world: A controllable
  generative world model for robot manipulation,'' \emph{arXiv preprint
  arXiv:2510.10125}, 2025.

\bibitem{guo2026vlaw}
Y.~Guo, T.~Lee, L.~X. Shi, J.~Chen, P.~Liang, and C.~Finn, ``Vlaw: Iterative
  co-improvement of vision-language-action policy and world model,''
  \emph{arXiv preprint arXiv:2602.12063}, 2026.

\bibitem{kingma2013auto}
D.~P. Kingma and M.~Welling, ``Auto-encoding variational bayes,'' \emph{arXiv
  preprint arXiv:1312.6114}, 2013.

\bibitem{peebles2023scalable}
W.~Peebles and S.~Xie, ``Scalable diffusion models with transformers,'' in
  \emph{Proceedings of the IEEE/CVF international conference on computer
  vision}, 2023, pp. 4195--4205.

\bibitem{perez2018film}
E.~Perez, F.~Strub, H.~De~Vries, V.~Dumoulin, and A.~Courville, ``Film: Visual
  reasoning with a general conditioning layer,'' in \emph{Proceedings of the
  AAAI conference on artificial intelligence}, vol.~32, no.~1, 2018.

\bibitem{liu2025factr}
J.~J. Liu, Y.~Li, K.~Shaw, T.~Tao, R.~Salakhutdinov, and D.~Pathak, ``Factr:
  Force-attending curriculum training for contact-rich policy learning,''
  \emph{arXiv preprint arXiv:2502.17432}, 2025.

\bibitem{intelligence2025pi}
P.~Intelligence, K.~Black, N.~Brown, J.~Darpinian, K.~Dhabalia, D.~Driess,
  A.~Esmail, M.~Equi, C.~Finn, N.~Fusai, \emph{et~al.}, ``{$\pi_{0.5}$}: A
  vision-language-action model with open-world generalization,'' \emph{arXiv
  preprint arXiv:2504.16054}, 2025.

\bibitem{zhao2023learning}
T.~Z. Zhao, V.~Kumar, S.~Levine, and C.~Finn, ``Learning fine-grained bimanual
  manipulation with low-cost hardware,'' \emph{arXiv preprint
  arXiv:2304.13705}, 2023.

\end{thebibliography}

\end{document}